\pdfoutput=1

\documentclass[11pt]{article}

\usepackage{ACL2023}

\usepackage{times}
\usepackage{latexsym}

\usepackage[T1]{fontenc}

\usepackage[utf8]{inputenc}
\usepackage{graphicx}
\usepackage{amsfonts}
\usepackage{booktabs}
\usepackage{microtype}

\usepackage{inconsolata}
\usepackage{multirow}
\usepackage{microtype}
\usepackage{graphicx}
\usepackage{subfigure}
\usepackage{booktabs} 
\usepackage{amsfonts}
\usepackage[utf8]{inputenc} 
\usepackage[T1]{fontenc}    
\usepackage{url}            
\usepackage{booktabs}       
\usepackage{amsfonts}       
\usepackage{nicefrac}       
\usepackage{microtype}      
\usepackage{xcolor}         
\usepackage{caption}		
\usepackage{wrapfig}		
\usepackage{multirow}		
\usepackage{booktabs,arydshln}
\usepackage{threeparttable}
\usepackage{tikz}

\makeatletter
\def\adl@drawiv#1#2#3{%
        \hskip.5\tabcolsep
        \xleaders#3{#2.5\@tempdimb #1{1}#2.5\@tempdimb}%
                #2\z@ plus1fil minus1fil\relax
        \hskip.5\tabcolsep}
\newcommand{\cdashlinelr}[1]{%
  \noalign{\vskip\aboverulesep
           \global\let\@dashdrawstore\adl@draw
           \global\let\adl@draw\adl@drawiv}
  \cdashline{#1}
  \noalign{\global\let\adl@draw\@dashdrawstore
           \vskip\belowrulesep}}
\makeatother

\title{pNLP-Mixer: an Efficient all-MLP Architecture for Language}

\author{Francesco Fusco\\
IBM Research \\
\texttt{ffu@zurich.ibm.com} \\\And
Damian Pascual\thanks{\ \ Work done during a research stay at IBM Research.}\\
Telepathy Labs\\
\texttt{damian.pascual@telepathy.ai} \\\AND
Peter Staar\\
IBM Research\\
\texttt{taa@zurich.ibm.com} \\\And
Diego Antognini\\
IBM Research\\
\texttt{Diego.Antognini@ibm.com} \\}
\begin{document}

\maketitle

\begin{abstract}

Large pre-trained language models based on transformer architecture
have drastically changed the natural language processing~(NLP)
landscape.~However, deploying those models for~on-device applications in
constrained devices such as smart watches is completely impractical due to
their size and inference cost. As~an~alternative to transformer-based
architectures, recent work on efficient NLP has shown that weight-efficient
models can attain competitive performance for simple tasks, such as slot
filling and intent classification, with model sizes in the order of the
\textit{megabyte}.~This work introduces the pNLP-Mixer architecture, an
embedding-free MLP-Mixer model for on-device NLP that achieves high weight-efficiency
thanks to a \textit{novel projection layer}. We evaluate a pNLP-Mixer model of only
\textit{one megabyte} in size on two multi-lingual semantic parsing
datasets, MTOP and multiATIS. Our quantized model achieves
99.4\% and 97.8\% the performance of mBERT on MTOP and multiATIS, while using \textit{170x fewer parameters}.
Our model consistently beats the~state-of-the-art
of~tiny models~(pQRNN), which is twice as large, by a margin up to 7.8\% on MTOP.

\end{abstract}

\section{Introduction}

Large language models based on transformer architecture have been
fueling the latest~successes in natural language processing~(NLP). Nowadays,
fine-tuning pre-trained models represents the de-facto framework to tackle
diverse NLP tasks, even those with limited amounts of annotations.

While the merit of large pre-trained language models is undeniable, using
models of several gigabytes and billions of parameters is not always practical
or even possible due to computational and memory requirements. In
addition, there are many simple and yet important tasks, such as slot filling in home assistants, which do not require the complex linguistic knowledge
encoded by large pre-trained models and for which smaller models may reach
competitive performance at a much lower cost. Reducing model sizes to the order
of the \textit{megabyte} is a necessity for resource constrained devices, such
as smart watches, and in general it is attractive for edge use cases
as i) updating models at the edge requires pushing updates to potentially
millions of devices, and ii) multiple models solving different tasks can be
deployed even on embedded devices with limited memory~capacity.

Transformer-based architectures are not suitable to downscale to such \textit{ultra
small} model sizes, mostly due to the space required to store embedding tables~\cite{zhao-etal-2021-extremely}. Projection-based models~\cite{DBLP:journals/corr/abs-1708-00630} have shown
that the dense~representations learned as part of the training process and
stored in the embedding tables can be replaced by non-trainable
representations computed on-the-fly over the text, hence the name
embedding-free.

In this work, we introduce the \textit{pNLP-Mixer}, a novel embedding-free
architecture for ultra-small NLP models targeting on-device applications. Our
architecture relies on a novel \textit{projection layer} which creates text
representations for individual tokens by combining the MinHash
fingerprints~\cite{Broder00identifyingand} corresponding to each subword unit.
The projected features are given as input to a
MLP-Mixer~\cite{tolstikhin2021mixer}, which grants our model architecture
linear scalability in the sequence length and seamless hardware acceleration.
To the best of our knowledge, this is the first work combining subword-unit
tokenization and MinHash fingerprints in projection networks.

Our evaluation on two semantic parsing datasets representative of on-device
applications, MTOP and multiATIS, showcases that the pNLP-Mixer beats the
current state-of-the-art for ultra-small models,
pQRNN~\cite{DBLP:journals/corr/abs-2101-08890}, by up to 7.8\% on sequence
tagging tasks. On MTOP, a pNLP-Mixer model with only one million parameters
achieves 99.4\% of the performance of mBERT, which has \textit{170x} more
parameters.

\section{Related Work}

Since the introduction of transformer-based language models such as
BERT~\cite{devlin-etal-2019-bert}, model sizes have been increasing at
unprecedented pace~\cite{NEURIPS2020_1457c0d6,
goyal-etal-2021-larger,lample2019cross}. Using current large language models
for on-device applications is simply not feasible due to the size and
computational requirements, especially in resource constrained devices such as
smart watches. Transformer-based models optimized for smartphone use cases,
such as DistilBERT~\cite{DBLP:journals/corr/abs-1910-01108},
TinyBERT~\cite{jiao-etal-2020-tinybert}, and
MobileBERT~\cite{sun-etal-2020-mobilebert} have shown that by combining
knowledge distillation~\cite{44873} and quantization~\cite{Jacob_2018_CVPR},
one can achieve model sizes in the order of tens to hundreds of megabytes in
size. Embedded devices, such as wearables, require instead model sizes in the
order of the \textit{megabyte}, a target that is very challenging to achieve
with transformer-based architecture, mostly because of the size of the
embedding tables~\cite{zhao-etal-2021-extremely}.

Embedding-free model architectures have been introduced to completely eliminate
the dependency on large embedding tables from models. Instead of learning
embeddings at training time, text representation are computed on-the-fly using
solely the surface forms of the tokens by means of locality-sensitive
hashing~(LSH)~\cite{charikar2002similarity}~techniques. This way tokens that
are similar at the surface level have similar representations. The idea of
replacing trainable parameters stored in embedding tables with LSH-based
projections has been introduced in \citet{DBLP:journals/corr/abs-1708-00630} and
\citet{pmlr-v97-ravi19a}. Follow up research work on model architectures
targeting ultra-small model sizes has resulted in several model architectures
including
SGNN~\cite{ravi-kozareva-2018-self-governing}, SGNN++~\cite{pmlr-v97-ravi19a},
Prado~\cite{kaliamoorthi-etal-2019-prado}, and
pQRNN~\cite{DBLP:journals/corr/abs-2101-08890}. Our model architecture belongs
to the same line of research, but introduces a linguistically informed
projection layer which combines subword-unit tokenization~\cite{sennrich2016}
with LSH principles. In our work, we evaluate and compare multiple LSH
techniques, including SimHash~\cite{Manku2007} and
MinHash~\cite{Broder00identifyingand}. In our projection layer, by exploiting
the associativity property of MinHash, fingerprints of individual tokens can be
efficiently computed from the fingerprints of their subword units.

Our model does not use attention mechanisms, as it feeds
the representations to a MLP-Mixer~\cite{tolstikhin2021mixer} model.
While using MLP only architectures is not new in the NLP landscape
\cite{NEURIPS2021_4cc05b35,arxiv.2203.06850}, this work is the first proposing
an all-MLP architecture for ultra-small models. There are numerous studies 
around efficient transformer-based models~\cite{10.1145/3530811} and
solutions to make them scale linearly with the sequence length. However, none
of those work targets models of the size of the single megabyte.

\section{pNLP-Mixer: a Projection MLP-Mixer}
\label{sec:model}

The pNLP-Mixer has been designed from the ground up as an efficient architecture
suitable for both edge cases, memory and latency constrained, and as a 
backbone for complex NLP pipelines.

\begin{figure}[t]
\begin{center}
      \includegraphics[width=0.95\linewidth]{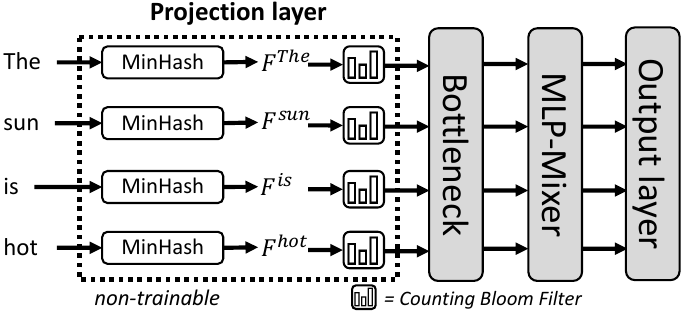}
\end{center}

\caption{\label{fig:architecture}
    Our pNLP-Mixer model has a \textit{non-trainable} projection layer which feeds
    a MLP-Mixer architecture with rich text features representing MinHash \textit{fingerprints} in a Counting Bloom Filter.}

\end{figure}

Figure~\ref{fig:architecture} depicts the model architecture at high level. The pNLP-Mixer
falls into the category of projection-based models: instead of storing large
embedding tables, like transformer-based models do, our model uses a projection
layer which captures morphological knowledge from individual tokens using
\textit{non-trainable} hash functions.
This projection layer can be seen as a feature
extractor that produces a representation from input text. Once the input
features are computed, they are passed through a trainable linear
layer called bottleneck layer. The output of the bottleneck
layer is the input of a series of MLP blocks of a standard MLP-Mixer
architecture~\cite{tolstikhin2021mixer}.

There are several advantages of using an all-MLP architecture for language
processing. In contrast to attention-based models, the MLP-Mixer captures long-range dependencies without introducing a quadratic cost on the sequence
length. Further, by using only MLPs, the model becomes simple to implement and has out-of-the-box hardware acceleration in devices ranging from mobile
phones to server-grade inferencing accelerators.

The main contribution of our work is to show that a simple model like the
MLP-Mixer represents a valid alternative to transformer-based models in
NLP, even in setups where large embedding tables are replaced with projections computed on the fly.
The key to achieve competitive performance with such
small and computationally efficient models is to feed them with high-quality input features. 

\subsection{Projection Layer}
\label{sec:model:projection}

Our projection layer builds upon the notion of locality sensitive
hashing~(LSH)~\cite{indyk1998approximate} to create representations from text.
While LSH has been introduced in previous works, e.g., in
pQRNN~\cite{DBLP:journals/corr/abs-2101-08890}, our approach is completely
novel. In particular, we combine subword-unit
tokenization~\cite{37842,sennrich2016} and
the \textit{associativity} of MinHash~\cite{Broder00identifyingand} to
efficiently compute features of any token as a combination of the features
corresponding to its subword unit. Subword tokenization, which is commonly used
in transformers, ensures that any text can be represented as a sequence of
subwords units, i.e., there are no out-of-vocabulary words. In our context, using
subword tokenization provides two main advantages: i) linguistic
knowledge can be injected by training domain-specific subword-unit tokenizers,
and ii) the representation of each subword unit can be precomputed and cached to
reduce inference costs. 

\begin{figure}[t]
\begin{center}
      \includegraphics[width=.775\linewidth,height=3.2cm]{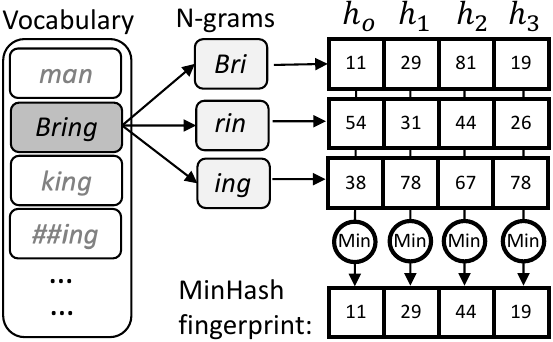}
\end{center}
\caption{\label{fig:fingerprint}The MinHash fingerprint of a subword unit contains the minimum hash
    values computed over the trigrams, for each hash function $H_{0-3}$. Fingerprints for
    a given token are computed by aggregating the fingerprints of its subword units in a similar way.}
\end{figure}

Our projection layer calculates the MinHash fingerprint $F^{t}$ of each input
token $t$ by reusing the fingerprint of individual subword units
belonging to the vocabulary $V$ (see Figure~\ref{fig:fingerprint}). A fingerprint $F \in \mathbb{N}^n$ is an array of $n$
positive integers $F_{0}$ to $F_{n-1}$, computed with $n$ distinct hash
functions $h_0(x)$ to $h_{n-1}(x)$ mapping strings to positive integers.
This way, the first step of our projection is tokenization, which transforms each input token into
a list of subword units. Then, for
each subword unit $u$, we calculate its fingerprint $F^{u}$.
Each element $F^{u}_i$, with $0 \leq i < n$,
is obtained by first applying a hash function
$h_i(x)$ to each of the trigrams $v_{0}$ to $v_{k-1}$ extracted from
the subword $u$, with $k \geq 1$. 
Then, $F^{u}_i$ is obtained as the minimum hash value across trigrams: $F^{u}_i=min(h_i(v_0),...,h_i(v_{k-1}))$.~For example, for the subword unit $``Bring"$, $F^{Bring}_{i}$ is computed as 
$F^{Bring}_{i}$=$min($$h_{i}(``Bri"),h_{i}(``rin")$,$h_{i}(``ing"))$.

When a subword is a continuation, e.g., $``\#\#ing"$, we skip the trigram extraction and 
calculate the hash $h_i(x)$ directly on the full subword unit $u$. The fingerprint
$F^{u}$ is built by calculating $F^{u}_i$ for each of the $n$ hash functions $h_0(x)$ to $h_{n-1}(x)$.

Finally, the fingerprint $F^{t}$ of a token $t$ made of several subword units
$u_{0}$ to $u_{j-1}$, e.g., $``Bringing"$ $\rightarrow [``Bring"$, $``\#\#ing"]$,
is simply obtained by setting each element $F^{t}_i$ to the minimum across the
$j$ subword fingerprints $F^{u_0}_i$ to $F^{u_{j-1}}_i$. In our example,
$F^{Bringing}_{i} = min(F^{Bring}_i, F^{\#\#ing}_i)$.

In practice, if the fingerprint of each subword unit $u$ in the vocabulary $V$
is precomputed and cached, inference does not require \textit{any hashing on
strings} but only computing the \textit{minimum} between integer sets. In our
setup, we use the minimum operator as described in the original MinHash
paper~\cite{Broder00identifyingand}, which also contains the required
theoretical foundations. What we introduce in our work is a method that
elegantly exploits the associativity property of MinHash to avoid computing hash
functions over strings at runtime.

For each input token $t$, we do not use the fingerprints directly as input to the
bottleneck layer but, instead, we use them to populate an
array $C^{t} \in \mathbb{R}^m$ of $m$ float counters initially set to zero. In detail,
given the fingerprint $F^{t}$ corresponding to the token $t$, for each of the $n$ MinHash 
values $F^{t}_{i}$, we increase by one the value in position $p$ of the array of 
counters $C^{t}$, where $p=F^{t}_{i}$ $mod$ $m$. Therefore, the extracted feature
from each token is essentially a Counting Bloom Filter~\cite{fan2000summary} storing the
set of integers part of its MinHash fingerprint.

Caching fingerprints for subword-units is entirely \textit{optional}. The
memory required to enable caching is given by an integer matrix storing $|V|$
fingerprints of size $n$, where $|V|$ is the vocabulary size and $n$ the number
of hash functions. In practice, caching costs just a few megabytes of memory
with vocabularies from pre-trained models and fingerprints with, e.g., $64$
hashes. Note, that there is a fundamental difference between the embedding
matrices stored in transformers and our cached fingerprints. In contrast to
embedding tables, which in transformer-based models are task specific as a
result of the fine-tuning process, the fingerprints are \textit{not trainable}
and they are not directly involved in any matrix multiplication. Since
fingerprints are not trainable, they can be reused across different models,
i.e., a \textit{single} cache can serve $n$ models. In complex NLP pipelines to
be executed on embedded devices at the edge, this architecture provides
substantial opportunities for optimizations. First, the same token fingerprint
can be reused to perform the inference with distinct models, which means that
the cost of computing the projection can be easily amortized. Second, as long as
tokenizer and hashing scheme do not change, distinct models can be
independently updated, while keeping the cache of subword-unit fingerprints
unmodified on the device. Third, having a cache that is shared among models
means that the memory costs required to enable caching are also amortized.
It is worth to remark that the advantages offered by our architecture are not
limited to edge use-cases. Large-scale natural language processing platforms
running in data-centers can equally benefit from the resource optimization and
granular deployment opportunities offered by our architecture.

\subsection{MLP-Mixer}

The MLP-Mixer~\cite{tolstikhin2021mixer} is a simple architecture that consists exclusively of mixer blocks. Each block has two multi-layer perceptrons (MLPs) interleaved by a transposition operation. 
The transposition of the output of the first MLP lets the second operate on the sequence dimension, effectively mixing information across tokens.
Our model follows the original work.

In our case, the matrix $C \in \mathbb{R}^{s \times m}$ produced by the projection layer, where $s$ the sequence length and $m$ the size of the counting bloom filter, is
passed through a bottleneck layer: a dense layer followed by an activation function and a normalization layer, that outputs a matrix 
$B \in \mathbb{R}^{s\times h}$, where $h$ is the hidden size.
$B$ is fed to the MLP-Mixer, which in turn produces an output $O \in \mathbb{R}^{s\times h}$. 
We~apply a classification head on top of $O$ to generate
the predictions. In the case of semantic parsing this head is a linear layer applied on each token, while for classification tasks, we use a max pooling instead.

\section{Experimental Setup}
\label{sec:expsetup}

Our architecture is designed as an alternative to transformer-based
models for \textit{ultra-small} models (i.e., one megabyte) targeting on-device applications. In the
league of extremely small models, common evaluation datasets used by research
and industry are not the same as the ones used for evaluating the generalizability of large pre-trained
language models~(e.g., GLUE \cite{wang-etal-2018-glue}),~but datasets for simpler tasks, that are more
realistic applications for tiny models. Thus, we align~to prior
works on tiny models for on-device applications \cite{kaliamoorthi-etal-2019-prado,DBLP:journals/corr/abs-2101-08890} that assess models on two multilingual semantic parsing datasets.

\paragraph{MTOP} \cite{mtop}. It covers six languages, English, Spanish, French, German, Hindi, and Thai. It was created by translating from English to the other languages. The train, dev,
and test~set for each language contain 10k,
1.5k, and 3k samples. 
We assess the models on 
slot parsing (\textit{named entity recognition}) on 78 different slot~labels. We report the exact match accuracy score, computed as the number of instances whose \textit{all} tokens have been correctly labeled over the number of instances.

\paragraph{multiATIS} \cite{multiatis}. It is a multilingual version of the ATIS dataset~\cite{atis}, that contains
queries related to air travel in nine languages: English, Spanish, French, German, Hindi, Japanese, Portuguese,
Turkish, and Chinese. Each language except Hindi and Turkish consists of $4,488/490/893$ samples for train, dev, and test sets; for Hindi and Turkish the splits are $1,440/160/893$ and $578/60/715$.
We evaluate on \textit{intent classification}: determining the intent of a query from 18 labels, and we report the accuracy.

\paragraph{Training Details.}~In our experiments, we aim for model sizes in the order of \textit{one million parameters}~(one megabyte with 8-bit quantization).
All trained models are approximately of this size. For the pNLP-Mixer, the projection of each token is a feature vector of dimension $512$ filled with $256$ hashes. The bottleneck consists of one MLP with a Leaky ReLU as the activation function~\cite{xu2015empirical} followed by a normalization layer~\cite{ba2016layer}. Finally, we use $5$ Mixer layers where each block contains $256$ hidden dimensions for the token-mixing, channel-mixing, and classification head. We use the tokenizer of BERT-base multilingual cased. We tune the learning rate, weight decay, and dropout with a batch size of $128$ and using early-stopping with a patience of $5$ epochs. 
We select the models reaching~the best exact match and intent accuracy on the validation set. We report their performance on the test~set.

\begin{table}[!t]
\centering
\small
\begin{tabular}[!t]{@{}lcc@{}}
\label{proj}
\centering
& MTOP EN & multiATIS EN\\
Projection & Exact Match Acc. & Intent Acc. \\
\midrule
Binary      & $80.58$ & $97.97$\\
TSP        & $80.33$ & $98.17$\\
SimHash    & $80.99$ & $98.38$\\
MinHash (Ours)    & $\textbf{82.51}$ & $\textbf{98.57}$\\

\end{tabular}
\caption{\label{proj}Comparison of different projection layers followed by the bottleneck and MLP-Mixer on the validation set of English MTOP and multiAtis.}
\end{table}

\section{Model Investigation}
\label{sec:model_inv}

We provide detailed insights on the impact of different projection layers and
other architectural components as well as a comparison to alternative architectures. 
We perform the experiments on the English variant of the MTOP and multiATIS datasets.

\subsection{Projection Comparison}

First, given the pNLP-Mixer model of Section~\ref{sec:expsetup} with input
features fixed to 512, we compare different feature extraction strategies.
Specifically:

\medskip
\noindent
\textbullet\ \textbf{Binary.} We compute $256$ hash values for each token. Given a
token and a bitmap of size $m=512$ set to zero,
for each hash value $h_{v}$, we set to $1$ the bit in
position $p=h_{v}$ $mod$ $m$ of the bitmap. The token feature is a float tensor storing the bitmap.

\medskip
\noindent
\textbullet\ \textbf{TSP.} For each token a 1024-bits hash is computed and then
represented as ternary feature of size 512 as
described in~\citet{kaliamoorthi-etal-2019-prado}.

\medskip
\noindent
\textbullet\ \textbf{MinHash.} Our projection layer (Section~\ref{sec:model:projection}).

\medskip
\noindent
\textbullet\ \textbf{SimHash.} We compute the hashes of subword units as in \textit{MinHash}, but
we combine them using SimHash~\cite{Manku2007,Shrivastava014}. The extracted feature is a binary
feature of size $l$, where $l$ is the size~(in bits) of the hashes applied to
n-grams or entire subword units. The value at index $0 \leq p < l$ of the
feature is the sign of the value $\phi_{p}$ stored at index $p$ of a histogram $\phi$ of length $l$.
The histogram, initialized to 0, is populated by summing or
subtracting $1$ to $\phi_{p}$ whenever a hash value has a $1$ or $0$ in position~$p$.

\medskip

In Table~\ref{proj}, we report the best scores obtained after tuning each
configuration. Overall, our projection layer MinHash obtains the best exact
match accuracy and intent accuracy, with an absolute improvement over SimHash
of $+1.52$ and $+0.19$. Binary and TSP obtain the worst performance: $-1.93$
and $-2.18$ on the MTOP compared to MinHash, and $-0.60$ and $-0.4$ on
multiATIS. Those differences confirm the limitation of binary and ternary features
and highlight the importance of carefully designing the
projection layer and justifies an effort for further research on projection
algorithms. Given these results, we only consider
our MinHash-based projection for the rest of the experiments.

\begin{table}[!t]
\small
\begin{tabular}{@{}l@{}c@{\hspace{3mm}}c@{\hspace{3mm}}c@{}}
\centering
& & MTOP EN & multiATIS EN\\
Model            & \# Param. & Exact Match Acc. & Intent Acc. \\
\midrule
Projection-only  & 0.2M   & $49.15$ & $81.54$\\
CNN              & 1.0M   & $73.74$ & $97.77$\\
LSTM             & 1.2M   & $76.92$ & $97.77$\\
Transformer      & 1.0M   & $74.05$ & $97.97$ \\
MLP-Mixer        & 1.0M   & $\textbf{82.51}$ & $\textbf{98.57}$\\
\end{tabular}
\caption{\label{tab:proj_vs_MLP}Comparison of different architectures using the MinHash projection layer on the validation set of English MTOP and multiATIS.}
\end{table}

\begin{table*}[!t]
\centering
\small
\begin{tabular}{@{}
l
l
c
c
c
c
c
c
c
c
c|
c@{}}
  & & \multicolumn{9}{c}{Intent Accuracy} & \\
 \cmidrule(lr){3-12}
Model             & \# Param.    & EN     & ES     & FR     & DE     & HI     & JA     & PT     & TR     & ZH     & Avg    \\
\midrule
LSTM              &  28M  & $96.1$ & $93.0$ & $94.7$ & $94.0$ & $84.5$ & $91.2$ & $92.7$ & $81.1$ & $92.5$ & $91.1$ \\
mBERT 
& 170M & \underline{$98.3$} & \underline{$97.4$} & \underline{$98.6$} & \underline{$98.5$} & \underline{$94.5$} & \underline{$98.6$} & \underline{$97.4$} & \underline{$91.2$} & \underline{$97.5$} & \underline{$96.9$}\\
\midrule
Transformer 
& 2M   & $96.8$ & $92.1$ & $93.1$ & $93.2$ & $79.6$ & $90.7$ & $92.1$ & $78.3$ & $88.1$ & $89.3$ \\
pQRNN 
& 2M\textsubscript{(8bit)} & $98.0$ & $97.0$ & $97.9$ & $96.6$ & $\textbf{90.7}$ & $88.7$ & $\textbf{97.2}$ & $86.2$ & $93.5$ & $94.0$ \\
\cdashlinelr{1-12}
pNLP-Mixer & \multirow{1}{*}{1M\textsubscript{(8bit)}} & $\textbf{98.1}$ & $\textbf{97.1}$ & $\textbf{98.1}$ & $\textbf{97.3}$ & $\textbf{90.7}$ & $\textbf{92.3}$ & $\textbf{97.2}$ & $\textbf{87.3}$ & $\textbf{95.1}$ & $\textbf{94.8}$\\
\end{tabular}
\caption{\label{matis-test}Intent accuracy across languages on the test sets of multiATIS. For each language we \underline{underline} the best overall
result and we mark in \textbf{bold} the best performance among the tiny models.}
\end{table*}
\begin{table}[!t]
\centering
\small
\begin{tabular}{@{}
l@{\hspace{1mm}}
l@{}
c@{\hspace{1.5mm}}
c@{\hspace{1.5mm}}
c@{\hspace{1.5mm}}
c@{\hspace{1.5mm}}
c@{\hspace{1.5mm}}
c@{\hspace{1.5mm}}|
c@{}}
 & &  \multicolumn{7}{c}{Exact Match Accuracy}\\
 \cmidrule(lr){3-9}
Model             & \#Param.    &   EN   &   ES   &   FR   &   DE   &   HI   &   TH   &   Avg  \\
\midrule
XLU 
&  70M & $78.2$ & $70.8$ & $68.9$ & $65.1$ & $62.6$ & $68.0$ & $68.9$ \\
XLM-R 
& 550M & \underline{85.3} & $81.6$ & $79.4$ & \underline{76.9} & \underline{76.8} & $73.8$ & \underline{79.0} \\
mBERT 
& 170M & $84.4$ & \underline{81.8} & \underline{79.7} & $76.5$ & $73.8$ & $72.0$ & $78.0$ \\
\midrule
Transformer 
& 2M   & $71.7$ & $68.2$ & $65.1$ & $64.1$ & $59.1$ & $48.4$ & $62.8$ \\
pQRNN 
& \multirow{2}{*}{2M\textsubscript{(8bit)}}    & $78.8$ & $75.1$ & $71.9$ & $68.2$ & $69.3$ & $68.4$ & $71.9$ \\
- distilled 
&  & $79.4$ & $75.4$ & $73.0$ & $68.6$ & $70.2$ & $69.5$ & $72.7$ \\
\cdashlinelr{1-9}
pNLP-Mixer & \multirow{1}{*}{1M\textsubscript{(8bit)}} & $\textbf{84.0}$ & $\textbf{78.3}$ & $\textbf{75.2}$ & $\textbf{76.9}$ & $\textbf{76.5}$ & \underline{$\textbf{74.1}$} & $\textbf{77.5}$\\
\end{tabular}
\caption{Exact match accuracy across languages on the test sets of MTOP. We \underline{underline} the best overall
result for each language and mark in \textbf{bold} the best performance among the tiny models.}
\label{mtop-test}
\end{table}

\subsection{Model Comparison}\label{sec:model_comp}
Now, we investigate whether the MLP-Mixer is the optimal architecture to process
this representation. First, we remove the MLP-Mixer and connect the output of the bottleneck layer to the classification heads (Projection-Only). Then, we replace the MLP-Mixer with three alternative architectures: 
a convolutional neural network (CNN)~\cite{lecun2015deep}, 
a long short-term memory recurrent neural network (LSTM)~\cite{lstm}, and a transformer~\cite{NIPS2017_3f5ee243}.

Table ~\ref{tab:proj_vs_MLP} shows that using the projection-layer directly as input to
the classification heads without a model in between, results in very poor performance.
From the alternative models, all perform significantly worse than the MLP-Mixer: $-8.77$, $-5.59$, and $-8.46$~in terms of exact match accuracy for the CNN, LSTM, and transformer models, respectively. This last result is remarkable: for the same number of parameters, the MLP-Mixer outperforms
the transformer while having a linear complexity on the input length instead of a quadratic one. Overall, 
the evaluation shows that the MLP-Mixer is weight-efficient for processing the projection output and reaching high performance.

\section{Evaluation}

Finally, we run a complete evaluation on
the test sets of MTOP and multiATIS. 
We~compare our pNLP-Mixer with three very
large models: XLU~\cite{DBLP:journals/corr/abs-1909-07009}, which is a bi-LSTM model with pretrained XLU embeddings, and two pretrained multilingual models: XLM-R~\cite{conneau-etal-2020-unsupervised} and multilingual BERT (mBERT)~\cite{devlin-etal-2019-bert}. We also include two small models: pQRNN~\cite{DBLP:journals/corr/abs-2101-08890} and a simple transformer using the same projection as pQRNN.
pQRNN is an embedding-free Quasi-RNN \cite{DBLP:conf/iclr/0002MXS17} model that shares the same philosophy of our proposed
pNLP-Mixer: a small and task-specific model that learns directly from the text.
For a fair comparison against pQRNN, we quantize our pNLP-Mixer models and report the performance
on the 8-bit version. Finally, we include pQRNN distilled with mBERT on MTOP (the original study did not distill pQRNN on multiATIS). 
The performance values of all the baselines are taken from~\citet{DBLP:journals/corr/abs-2101-08890}.

\medskip
\noindent
\textbf{MTOP.}
Table~\ref{mtop-test} shows that the large pre-trained models, XLM-R and mBERT, obtain the highest scores. Notably, from the
smaller alternatives, our pNLP-Mixer with only~1M parameters, 8-bit quantization and no pretraining,
i.e., \textit{680x smaller than mBERT}, reaches an average exact match accuracy only
$0.5$ and $1.5$ points~lower than mBERT and XLM-R. It even beats mBERT in the non-European languages.
With those results, the pNLP-Mixer beats a twice larger pQRNN model across all languages by $7.8\%$ in average.
It even beats a pQRNN model distilled from mBERT by $6.6\%$ in average.

\medskip
\noindent
\textbf{multiATIS.}
Table~\ref{matis-test} shows a similar trend compared to the MTOP dataset. On average, the pNLP-Mixer performs better than pQRNN while being twice as small.~Remarkably, the pNLP-Mixer significantly outperforms the transformer model and the larger LSTM. Moreover, it reaches $97.8\%$ of the performance of mBERT while being 680x smaller.

\paragraph{Discussion.}
The results show that the pNLP-Mixer represents a very competitive model for the 
settings where the maximum model size is limited due to either memory or latency requirements. 
Our pNLP-Mixer models, with only 1M parameters and a size of one megabyte when quantized, reaches competitive scores in both datasets compared to mBERT, which is a 680x larger model.
This represents an important step towards ultra-small models~for NLP. To put numbers in perspective, for the non-quantized pNLPN-Mixer model, the inference latency with batch size 1 on a \textit{single CPU core} is as little as $2.4$ms,\footnote{We report the average latency across 100 samples on a Xeon E5-2690v4 processor and a PyTorch runtime.} 
with the projection layer taking $0.4$ms. 
Finally, we could not compare pNLP-Mixer with pQRNN in terms of FLOPS or latency because the authors did not make the code available; we are unable to produce comparable predictive performance with our implementation of~pQRNN.

\section{Conclusion}

We introduce pNLP-Mixer, the first embedding-free model based on the MLP-Mixer
architecture. Our main contribution is an efficient and yet effective
projection layer which combines MinHash fingerprints and subword-unit
tokenization to create rich token representations.
Our evaluation shows that the pNLP-Mixer beats the state-of-the-art of
tiny NLP models, pQRRN, and offers sequence tagging performances that are up to
$7.8\%$ higher while using \textit{half} of the parameters.~The results are
remarkable: a pNLP-Mixer model of only \textit{1 million parameters} provides a
performance of $99.4\%$ and $97.8\%$ on MTOP and multiATIS, respectively, compared to mBERT
which is a a pre-trained model with \textit{170x} more parameters.
Our pNLP-Mixer model is simple to implement and accelerate, and
provides competitive performance even without pre-training or
distillation. Our work demonstrates the importance of projection methods
and embedding-free architectures to advance the field of ultra-small models.

\bibliographystyle{acl_natbib}

\appendix

\end{document}